\documentclass[11pt]{article}

\usepackage[final]{acl}
\usepackage{mathtools}
\usepackage{times}
\usepackage{latexsym}

\usepackage[T1]{fontenc}

\usepackage[utf8]{inputenc}

\usepackage{microtype}

\usepackage{inconsolata}

\usepackage{graphicx}
\usepackage{microtype}
\usepackage{hyperref}
\usepackage{url}
\usepackage{booktabs}
\usepackage{authblk}
\usepackage{hyperref}

\title{Learning to Rank Salient Content for Query-focused Summarization}

\author[ ]{\textbf{Sajad Sotudeh}}
\author[ ]{\textbf{Nazli Goharian}\vspace{-0.3cm}}
\affil[ ]{IR Lab, Department of Computer Science, Georgetown University}
\affil[ ]{\normalsize \textcolor{darkblue}{\texttt{\{sajad,nazli\}@ir.cs.georgetown.edu}}}

\definecolor{darkblue}{rgb}{0.0, 0.0, 0.5} 

\usepackage{xspace}
\usepackage{algpseudocode}
\usepackage{enumitem}
\usepackage{soul}
\usepackage{color,soul}
\usepackage{subcaption}
\usepackage{amsmath}
\usepackage{multirow}
\usepackage{tcolorbox}

\newcommand\rouge{\textsc{Rouge}\xspace}
\newcommand\bart{\textsc{Bart}\xspace}
\newcommand\segenc{\textsc{SegEnc}\xspace}

\newcommand\socratic{\textsc{Socratic}\xspace}
\newcommand\pegasus{\textsc{Pegasus}\xspace}
\newcommand\qontsum{\textsc{Qontsum}\xspace}
\newcommand\ours{\textsc{LTRSum}\xspace}

\begin{document}
\maketitle
\begin{abstract}
This study examines the potential of integrating Learning-to-Rank (LTR) with Query-focused Summarization (QFS) to enhance the summary relevance via content prioritization. Using a shared secondary decoder with the summarization decoder, we carry out the LTR task at the segment level. Compared to the state-of-the-art, our model outperforms on QMSum benchmark (all metrics) and matches on SQuALITY benchmark (2 metrics) as measured by Rouge and BertScore while offering a lower training overhead. Specifically, on the QMSum benchmark, our proposed system achieves improvements, particularly in Rouge-L (+0.42) and BertScore (+0.34), indicating enhanced understanding and relevance. While facing minor challenges in Rouge-1 and Rouge-2 scores on the SQuALITY benchmark, the model significantly excels in Rouge-L (+1.47), underscoring its capability to generate coherent summaries. Human evaluations emphasize the efficacy of our method in terms of relevance and faithfulness of the generated summaries, without sacrificing fluency. A deeper analysis reveals our model's superiority over the state-of-the-art for broad queries, as opposed to specific ones, from a qualitative standpoint. We further present an error analysis of our model, pinpointing challenges faced and suggesting potential directions for future research in this field.
\end{abstract}

\section{Introduction}
Query-focused summarization (QFS) is gaining prominence in research community. Unlike conventional summarization tasks that aim to capture the overall essence of a document or a set of documents, QFS focuses on generating concise summaries in response to {posed} queries. This specialization enables a more targeted information retrieval process, offering summaries that directly address the informational needs rather than providing a broad overview of the source material. 

The advancements in QFS have been notably driven by the introduction of invaluable datasets of long documents such as QMSum with an average of 9K tokens~\cite{Zhong2021DialogLMPM} and SQuALITY with an average of 5.2K tokens~\cite{wang-etal-2022-squality}, which have facilitated deeper exploration and innovation in this field. These datasets have laid the groundwork for the development of Transformers-based models which have shown strong potential in generating summaries that respond accurately to queries~\citep{Su2021ImproveQF, Laskar2022QFS, Vig2022SEGENC, Pagnoni2022SocraticPQ, Sotudeh2023QontSumOC,Yu2023ImprovingQM}. However, despite this proficiency, their ability to effectively \textit{prioritize information}—assessing its importance relative to a query to enhance summary relevance—remains an area for improvement.  {This study seeks to address this limitation, with a particular focus on long-input QFS, where summarizing multiple segments~\footnote{A chunk of document with a predefined length (e.g., 512 tokens).} for a given query presents unique challenges in capturing and prioritizing relevant content.}

Particularly, in this study, we present a novel enhancement to QFS through the incorporation of learning-to-rank (LTR), a technique with established efficacy in Information Retrieval. Our approach aims to refine the system's capability to discern and prioritize content segments not only by their relevance but also by their relative importance. This methodological advancement ensures that the produced summaries more accurately reflect the query's intent and hierarchically organize information by its significance. Central to our approach is the augmentation of use of the decoder that \textit{shares parameters} with the summarization decoder~\footnote{Particularly, we use the single decoder for two tasks: summarization and learning-to-rank.}, specifically designed for executing the LTR task at the segment level. {While Learning-to-Rank (LTR) is a classic approach, our innovation lies in adapting LTR principles specifically for Query-Focused Summarization (QFS) at the segment level, which has been less explored in the literature.} This strategy, inspired by the work of \citep{Zhuang2022RankT5FT} in adapting the T5~\citep{2020t5} encoder-decoder framework for text ranking in query-document scenarios, is tailored to address the nuances of segment ranking within the QFS context. Through the joint fine-tuning of summarization with cross-entropy loss, and LTR task—utilizing listwise cross-entropy softmax loss, our system not only elevates the relevance of generated summaries but also introduces a nuanced understanding of information importance. This strategy can aid the summarization system at attending to the source content given their relative importance. In short, our contributions are threefold: 
\begin{itemize}
    \item  We propose an LTR-assisted system for QFS that integrates the intuition of ranking and relative importance of segments during the summary generation process; 
    \item  Our proposed system outperforms across all automatic metrics (QMSum) and attains comparable performance in two metrics (SQuALITY) with lower training overhead compared to the SOTA. Additionally, our system enhances the relevance and faithfulness of generated summaries without sacrificing fluency; 
    \item We undertake an error analysis to discern the challenges faced by our model including label imbalance, and segment summarizer's hurdles, providing insights into potential avenues for further research.
\end{itemize}

\section{Related work}
The field of Query-focused Summarization (QFS)~\cite{Dang2005OverviewOD} has evolved significantly over the years, transitioning from early unsupervised extractive models \citep{Mohamed2006ImprovingQS,Wan2007ManifoldRankingBT,Zhao2010QueryfocusedSB,  Badrinath2011ImprovingQF, litvak-vanetik-2017-query} to recent approaches leveraging Transformer-based models \citep{Vaswani2017Att, Lewis2020BARTDS, zhang2019pegasus}. This evolution has been marked by the introduction of various techniques aimed at improving the relevance of summaries. Passage retrieval techniques \citep{Baumel2018QueryFA, Laskar2022QFS, Su2021ImproveQF, Zhong2021DialogLMPM, wang-etal-2022-squality}, transfer learning from the QA task~\cite{xu-lapata-2020-coarse, Zhang2021ScalingUQ, yuan-etal-2022-shot}, query modeling~\cite{Xu2020GeneratingQF,Xu2022DocumentSW,Yu2023ImprovingQM}, segment encoding \cite{Vig2022SEGENC}, and attention mechanisms to capture query-utterance relations \citep{Liu2023TUQFS} have all played a pivotal role in this advancement. Furthermore, the adoption of question-driven pretraining \citep{Pagnoni2022SocraticPQ} and contrastive learning \citep{Sotudeh2023QontSumOC} has introduced new dimensions to the task, simplifying the identification and summarization of salient content. More recently, \citet{liu-xu-2023-learning} {introduced the Ranker-Generator framework, which ranks utterances by learning from pairwise comparisons and global ordering. The top-ranked utterances are then used as input for the generator in summary generation.}

{While these methods have advanced QFS, existing techniques tend to treat all content segments equally without explicitly considering their relative importance within long-input  QFS tasks, which require processing large amounts of text and identifying key segments.} Hence, the comprehensive modeling of segment importance within the long QFS task remains a less explored area. Our approach introduces a novel application of learning-to-rank (LTR)\cite{Burges2005LearningTR,Cao2007LearningTR} mechanism to address this challenge, drawing inspiration from the successful application of LTR in broader Information Retrieval contexts\cite{Wang2022SimLMPW,Li2023LearningTR}.


\section{Background: Segment Summarizer (SegEnc)}
The backbone of current state-of-the-art systems for query-focused long summarization are built upon the Segment Encoding (\segenc) approach~\cite {Vig2022SEGENC}. \segenc operates by encoding fixed-length, overlapping segments of the source text, which are then integrated into a cohesive summary in an end-to-end manner, leveraging the decoder's ability to simultaneously attend to all encoded segments. To specifically adapt to query-focused summarization framework, \segenc embeds the query within each segment of the source text. This is achieved through a particular input framing technique, where the query is encapsulated by special markers and placed adjacent to each segment, adhering to the format: \texttt{<s>query</s>Segment}. This incorporation of the query into the summarization process is designed to enhance the focus on the query, offering a tailored approach to generating query-focused summaries.

\begin{figure*}[t]
    \centering
    \includegraphics[scale=0.58]{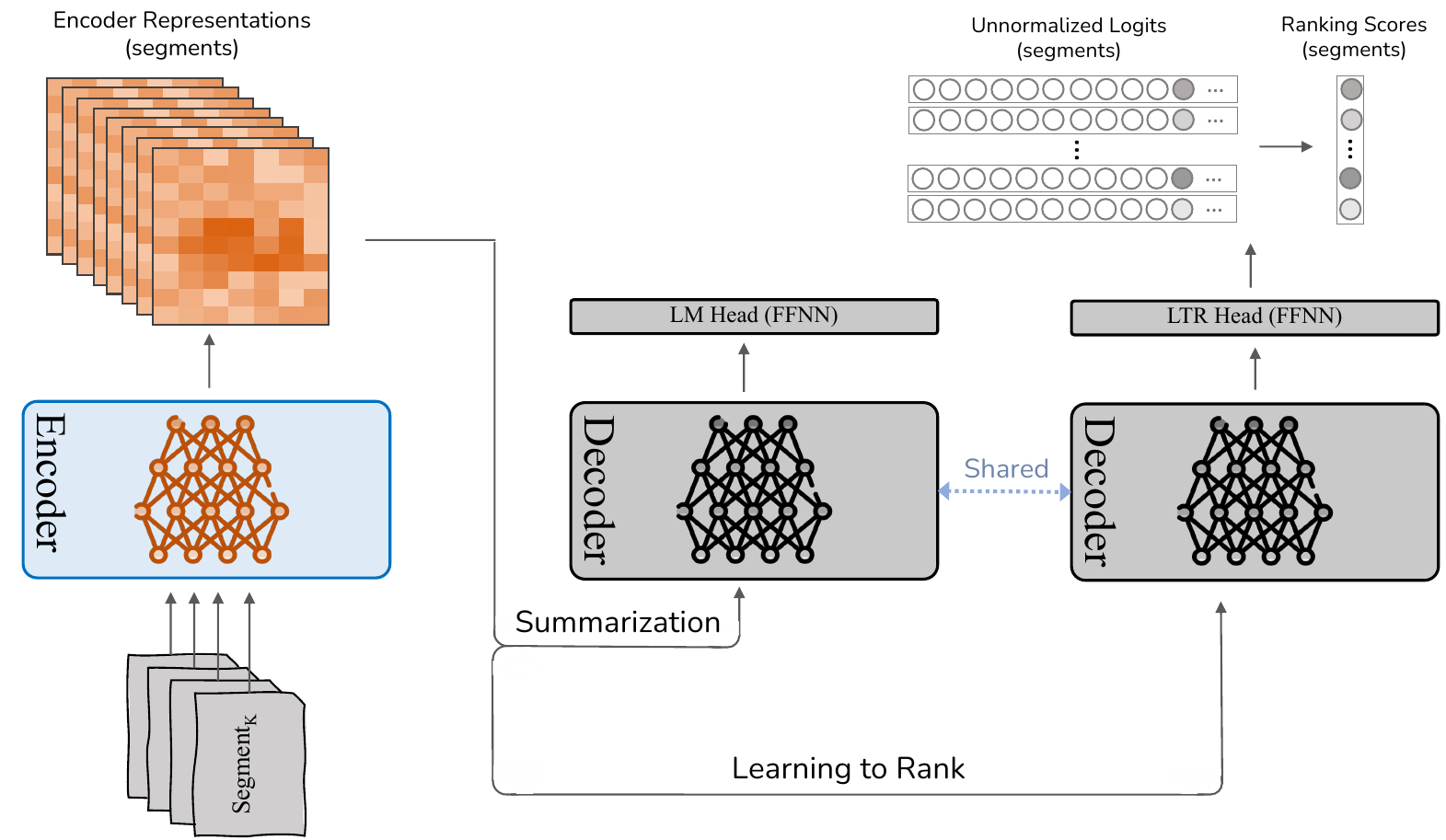}
    \caption{Overview of our proposed system (i.e., \textsc{LTRSum}). Our system utilizes a \textit{shared-parameter} decoder across two tasks, for the sake of learning to rank source segments (right-side decoder) alongside summarization (left-side decoder). It is important to note that our system uses a single decoder that shares parameters across both tasks, but for visual clarity, two decoders are depicted here. }
    \label{fig:model_overview}
\end{figure*}

\section{Model: LTR-assisted Summarization}

This study introduces an extension to the \segenc summarizer by integrating the Learning-to-Rank (LTR) principles, a notable ranking technique from the realm of information retrieval. This integration enables the summarizer to effectively learn the ranking of the gold segments. The segments' relevance labels are determined using a span probability-based heuristic (details in Section \ref{sec:exp_setup}) during the preprocessing step. An auxiliary LTR task is then formulated to instruct the summarizer in ranking source segments while performing the summarization task. 

{Figure} \ref{fig:model_overview} {shows the overview of our proposed system.  In particular, we exploit a \textit{shared} decoder to perform two tasks including summarization and learning-to-rank. This shared decoder operates by executing two forward passes, one for each task. For the LTR task,} following encoding of each segment (denoted as \texttt{Enc}(\texttt{S\textsubscript{i}})), \texttt{Dec\textsubscript{LTR}} takes in the segment encoder representations (i.e., the encoder representations associated with \texttt{<s>} token) and processes them through the LTR-dedicated decoder, followed by an LTR head (i.e., a feed-forward neural network (FFNN)) that is applied to the decoder outputs: 

\begin{equation*}
    \texttt{$\hat{y}$\textsubscript{i}} = \texttt{FFNN}(\texttt{Dec\textsubscript{LTR}}(\texttt{Enc}(\texttt{S}\textsubscript{i})))
\end{equation*}
wherein \texttt{S\textsubscript{i}} represents the i-th segment, and \texttt{$\hat{y}$\textsubscript{i}} corresponds to the decoder output for the same segment. Furthermore, an additional unused token is defined, analogous to the method described in \citep{Zhuang2022RankT5FT}, and its unnormalized logits are extracted from the decoder output $\hat{y}$\textsubscript{i} to serve as the segment ranking score: \texttt{rank\textsubscript{i}} = \texttt{$\hat{y}$\textsubscript{i <extra\_token\_id>}}. 

Having obtained the ranking outputs for all segments with the above procedure, a listwise softmax cross-entropy function is used to compute the Softmax loss as follows: 

\begin{equation*}
        \ell_{\texttt{Softmax}}(y_i, \hat{y}_i) = -\sum_{j=1}^{m} y_{ij} \log \left( \frac{e^{\hat{y}_{ij}}}{\sum_{j'=1}^{m} e^{\hat{y}_{ij'}}} \right)
\end{equation*}
where $y_i$ and $\hat{y_i}$ are the gold, and predicted relevance, respectively, and $m$ denotes the number of segments. After computing the Softmax loss, we combine it with the generation loss for joint training:

\begin{equation*}
         \ell_{\texttt{total}} = \ell_{\texttt{generation}} + 
\lambda \ell_{\texttt{Softmax}}(y_i, \hat{y}_i)
\end{equation*}
{in which $\ell_{\texttt{generation}}$ is a cross-entropy loss computed for generation task,} and $\lambda$ is a balancing parameter that should be tuned.

\section{Experimental Setup}
\subsection{Research questions }  {We seek to address the following research questions: 

\begin{itemize}
    \item \textbf{RQ1:} How does integrating the relative importance of segments into the summarization system influence the automatic and qualitative metrics of summaries?
    \item \textbf{RQ2: } How does the type of query (i.e., broad or specific) affect our system's performance compared to the SOTA?
    \item \textbf{RQ3: } What effect does the integration of LTR offer for segment retrieval?
    \item \textbf{RQ4: } What challenges does the model encounter in underperformed cases?
\end{itemize}

\label{sec:exp_setup}

\subsection{Datasets} We used two query-focused datasets during our study: (1) The QMSum dataset~\citep{zhong2021qmsum} consists of 1,808 query-focused summaries extracted from 232 multi-turn meetings across different domains. The dataset is split into training, validation, and testing sets with 1,257, 272, and 279 instances, respectively. The average source length is 9K tokens, and the summary length is 70 tokens. (2) SQuALITY~\citep{wang-etal-2022-squality} is a collection of question-focused abstractive summarization data with 100 stories, 500 questions, and 2,000 summaries. Each question is accompanied by four reference summaries written by trained writers. The dataset provides train/validation/test splits of 39/25/36, equivalent to 195/125/180 document-question pairs with average document and summary lengths of 5.2K and 237 tokens, respectively.


\subsection{Relevance labeling} Given the absence of relevance labels within the instances of datasets employed for experiments, we develop a probability-based heuristic to create such pseudo labels, which signifies the extent to which a segment aligns with the gold summary. Initially, the \textsc{SuperPal} approach, as mentioned in \citep{ernst-etal-2021-summary}, was employed as an external pseduo-labeling heuristic to match summary spans and their originating source spans, represented by a probability value, $p$. A specified threshold was then empirically determined for $p$, allowing only spans exceeding this threshold to be considered as gold during the labeling process. {Compared to other common approaches like greedy labeling}~\cite{Liu2019TextSW}{, which provides binary relevance labels based solely on hard matching criteria, our heuristic enables a probabilistic view that captures varying degrees of alignment. This scoring system, represented by the following equation, allows for a more continuous ranking of segments based on their relevance:}

\begin{equation*}
    \texttt{Score}(S_i) = \sum_j p_j \log(|\texttt{span}_j|)
\end{equation*}
where $S_i$ denotes the i-th segment, $\text{span}_j$ represents the j-th span within the segment $S_i$, {and $p_j$ shows the probability of $span_j$ being aligned to the gold summary.} {Intuitively, segments that have more common tokens with the gold summary (i.e., $|\texttt{span}_j|$) and assigned a higher probability by \textsc{SuperPal} approach (i.e., $p$), are more likely to be ranked higher.} Following the calculation of segment scores, they were organized in a sequence, and relevance labels were assigned according to the sorted scores.

{Our internal analysis found this heuristic to produce relevance labels that better align with summary content when compared to binary methods like greedy labeling. However, we acknowledge that alternative labeling strategies could be explored further in future research. The chosen approach ensures that segments most relevant to the query are ranked and labeled optimally within the context of our LTR techniques.}

\subsection{Implementation details } 
We built upon the code base provided by ~\citet{Vig2022SEGENC}, adhering to the default hyperparameters. The $\lambda$ hyperparameter was explored within the set \{0.5, 1, 1.5\}, and finally tuned to 1. Furthermore, a probability threshold ($p$) of 40\% was employed to filter gold segments. It has to be mentioned that all parameters, including $\lambda$ and $p$,  were empirically determined and fixed. Our model comprises 406 million parameters. We employed a single NVIDIA A6000 GPU for both training and evaluation. Each experimental training session spanned a duration of two days.

{\subsection{Comparison} {We compare our model to the well-established SOTA baselines on QFS:  }
\begin{itemize}[leftmargin=*,label={-}]
\item \textbf{{Ranker-Generator}}~\citep{liu-xu-2023-learning}: A recent abstractive summarizer that learns to rannk utterances from their relative orders, and then passes top-k utterances to the generator.

\item \textbf{\textsc{SegEnc}}~\citep{Vig2022SEGENC}: An abstractive summarizer that segments input, encodes and then decodes with joint attention. Versions include: (1) Finetuned on \bart large (\segenc); (2) pre-finetuned on Wikisum (\segenc-W);
\item \textbf{\textsc{Socratic}}~\citep{Pagnoni2022SocraticPQ}: A question-driven pre-training framework for controllable summarization, fine-tuned on \segenc. Also, a \pegasus variant pre-trained on \textit{Book3} is presented. 
\item \textbf{\qontsum}~\citep{Sotudeh2023QontSumOC}: A contrastive learning-based summarizer that distinguishes salient content from top-scored non-salient content. 
\end{itemize}

\begin{table*}[t]
  \begin{subtable}{1\linewidth}
    \centering
    \scalebox{1}{
    \begin{tabular}{lcccc}
\toprule
 & \textbf{RG-1} & \textbf{RG-2} & \textbf{RG-L} & \textbf{BS} \\
\midrule
Ranker-Generator~\cite{liu-xu-2023-learning} & 35.51 & 12.23 & 31.28 & - \\
\segenc~\citep{Vig2022SEGENC} & 37.05 & 13.03 & 32.62 & 87.44 \\
\hspace{.1em} \textit{\small+ Wikisum Pre-Finetuned}~\citep{Vig2022SEGENC} \normalsize& 37.80 & 13.43 & 33.38 & - \\
 \socratic Pret. 1M~\citep{Pagnoni2022SocraticPQ} \normalsize & 37.46 & 13.32 & 32.79 & 87.54 \\
\socratic Pret. 30M~\citep{Pagnoni2022SocraticPQ} \normalsize & 38.06 & {13.74} & 33.51 & 87.63 \\
\qontsum \citep{Sotudeh2023QontSumOC} &38.42 & {13.50} & {34.03} & 87.72 \\

\midrule
\ours (this work) & \textbf{38.82} & \textbf{14.11} & \textbf{34.45} & \textbf{88.07} \\

\bottomrule
\end{tabular}
    }
    \caption{}
    \label{subtab:first_table}
  \end{subtable}
  \\
  \\
  \begin{subtable}{1\linewidth}
    \centering
    \scalebox{1}{
    \begin{tabular}{lcccc}
\toprule
 & \textbf{RG-1} & \textbf{RG-2} & \textbf{RG-L} & \textbf{BS}\\
\midrule
\segenc~\citep{Vig2022SEGENC} & 45.68 & 14.51 & 22.47 & 85.86 \\
\hspace{.1em} \textit{\small+ Wikisum Pre-Finetuned}~\citep{Vig2022SEGENC} \normalsize& 45.79 & 14.53 & 22.68 & 85.96\\

 \pegasus Pret. ~\citep{Pagnoni2022SocraticPQ} & 45.78 & 14.43 & 22.90 & 85.94 \\
\socratic Pret. 30M~\citep{Pagnoni2022SocraticPQ} & \textbf{46.31} & \textbf{14.80} & 22.76 & 86.04 \\

\qontsum \citep{Sotudeh2023QontSumOC} & 45.76 & 14.27 & {24.14} & \textbf{86.07} \\

\midrule
\ours (this work) & 46.11 & 14.68 & \textbf{24.23} & {86.04} \\
\bottomrule
\end{tabular}
    }
    \caption{}
    \label{subtab:second_table}
  \end{subtable}
  \caption{Average of \rouge and \textsc{BertScore (BS)} performance of summarization baselines over (a) QMSum and (b) SQuALITY benchmarks. The baseline performances are reported from previous works. 
  }
  \label{tab:main_table}
\end{table*}

\begin{table}[ht]
\centering
\begin{tabular}{lcc}
\toprule
\textbf{Model}    & \textbf{QMSum} & \textbf{SQuALITY} \\ \midrule
\segenc          & 78.67                 & 253.12                  \\
\segenc-W           & 79.78                 & 245.53                  \\ 

\socratic          & 78.89                 & 241.23                  \\
\qontsum           & 77.45                 & 229.54                  \\
\ours            & 79.92                 & 226.76                  \\
\bottomrule
\end{tabular}
\caption{Average summary length for different models on QMSum and SQuALITY datasets}
\label{tab:summary_length}
\end{table}

\section{Experimental Results}

In this section, we present the automatic and human study results, followed by relevant analyses over query type impact, and segment retrieval.

\subsection{Automatic evaluation}
As shown in Table \ref{tab:main_table}, we compare the performance of our proposed system with existing state-of-the-art summarization techniques on the QMSum and SQuALITY benchmarks, employing 
\rouge and \textsc{BertScore} evaluation metrics to address RQ1 on automatic performance. For the QMSum benchmark, \ours surpasses state-of-the-art approaches. In particular, when compared with the \qontsum, our method achieves relative improvements of approximately 1.0\%, 4.5\%, 1.2\%, on the 
\rouge-1,
\rouge-2, 
\rouge-L metrics, respectively. Likewise, \ours surpasses \socratic Pret. by relative improvements of 2.0\% (\rouge-1), 2.7\% (\rouge-2), 2.8\% (\rouge-L. Additionally, the \textsc{BertScore} for \ours slightly edges out both \qontsum and \socratic Pret. 

On the SQuALITY dataset, \ours's performance reveals mixed results; over the \qontsum model, it slightly improves \rouge-1 and \rouge-2 metrics. However, when compared to \socratic Pret., \ours matches on \rouge-1 and \rouge-2 (with relative deficits under 0.01\%), demonstrates a remarkable 5.4\% improvement in 
\rouge-L and aligns closely with the \textsc{BertScore} metrics, on SQuALITY benchmark.  This is likely due to the challenges in automatically identifying high-quality ground-truth labels in SQuALITY, unlike QMSum, where our system benefits from human-annotated span labels, while the SQuALITY span labels were determined via a heuristic approach. Furthermore, another likely explanation for \socratic's performance boost may be attributed to its pretraining on the \textsc{Book3} dataset, which likely shares closer linguistic characteristics with the SQuALITY dataset.

It is essential to note that \socratic undergoes a large-scale pre-training process, driven by questions, which encompasses a vast number of examples drawn from the \textsc{Book3} corpus, amounting uo to 30M pre-training instances. This approach, while effective, is likely resource-intensive. Conversely, our model, \ours, bypasses the extensive pre-training stage and centers on learning an auxiliary task during the fine-tuning phase, making it a more resource-efficient alternative. {The \textsc{Rouge-L} improvement for QontSum and LTRSum, specifically on SQuALITY, is linked to their ability to generate concise summaries by focusing on key segments and reducing redundancy. As shown by the average summary lengths in Table} \ref{tab:summary_length}, {these models produce shorter summaries, which likely avoid unnecessary details and therefore improve alignment with reference summaries, leading to better ROUGE-L performance. In contrast, longer summaries from models like SegEnc may dilute relevance with introducing extraneous irrelevant information.}

\subsection{Ablation study}
{The effectiveness of sharing the decoder between the summarization and LTR tasks is demonstrated through the comparison between} \segenc and \ours models,{ presented in} Table~\ref{tab:main_table}. Specifically, the \segenc {model (which does not share the decoder) with our} \ours model (which uses a shared decoder). As shown in Table \ref{tab:main_table}, the \segenc {model serves as the vanilla baseline in our ablation, and the performance gains of} \ours {over this baseline highlight the contribution of the shared decoder used for the LTR and summarization tasks. Specifically, the shared decoder allows the model to leverage information from both the summarization and LTR tasks, leading to improvement gains across both the QMSum and SQuALITY datasets. This confirms that the architectural choice of a shared decoder is a key factor driving the performance improvements observed in our experiments.}




\subsection{Human evaluation} {We conducted human evaluations to assess the quality of the summaries generated by} \ours,  {in comparison with} \qontsum {and }\socratic~ {baseline systems. The evaluations were performed on the QMSum and SQuALITY benchmarks. Specifically, we randomly selected 64 test cases (QMSum) and 36 cases (entire test set of SQuALITY), resulting in a total of 100 cases. For each case, we provided two annotators}~\footnote{{Annotators were PhD students in Science and Engineering.}} with shuffled summaries, including the gold-spans from the source. To prevent bias, we shuffled summaries such that the correspondence could not be guessed. 
We then ask the annotators to score each case on a scale of 1 to 5 (worst to best) in terms of three qualitative metrics listed below, consistent with the ones employed by \citet{Sotudeh2023QontSumOC}: 
     \textbf{Fluency:} To gauge the understandability of a summary, {focusing on grammaticality, non-redundancy, and coherence aspects;} 
    \textbf{Relevance:} To assess the extent to which a summary is pertinent as an answer to the given query;
     \textbf{Faithfulness:} To measure the degree to which the content covered in the source is faithfully reflected in the generated summary.

\begin{table}[t]
\centering
\label{tab:human-eval}
\scalebox{0.85}{
\begin{tabular}{@{}lccc@{}}
\toprule
\multicolumn{1}{l}{\textbf{}} & \textbf{Fluency} & \textbf{Relevance} & \textbf{Faithfulness} \\ \midrule

\multicolumn{4}{l}{\textit{QMSum}} \\

\hspace{0.7cm}  \qontsum& 4.09 & {4.03} & {3.60} \\ 
\hspace{0.7cm} \socratic & {4.10}	 & {4.15} & {3.72} \\
\hspace{0.7cm} \ours & \textbf{4.14}	 & \textbf{4.36} & \textbf{3.88} \\

\midrule

\multicolumn{4}{l}{\textit{SQuALITY}} \\

\hspace{0.7cm}  \qontsum& {4.01} & {3.58} & 3.62 \\ 
\hspace{0.7cm} \socratic & \textbf{4.02}	 & {3.70} & {3.69} \\
\hspace{0.7cm} \ours & \textbf{4.02} & \textbf{3.81} & \textbf{3.78} \\

\bottomrule
\end{tabular}
}
\caption{Results of the human study on evaluation samples from the QMSum and SQuALITY datasets (64 cases from QMSum and 36 cases from SQuALTIY)}

\label{tab:human}
\end{table}
\begin{table*}[t]
  \begin{subtable}{1\linewidth}
    \centering
    \scalebox{1}{
    \begin{tabular}{llcccc}
    \toprule
        \textbf{Dataset} & \textbf{Query type (\%)} & \textbf{Flu.} &\textbf{ Rel.} & \textbf{Faith.} \\ 
        \midrule
         \multirow{2}{*}{QMSum} & Broad (53\%) & \textbf{29}/45/26 & \textbf{28}/55/17 & \textbf{25}/60/15 \\
         & Specific (47\%) & 21/54/\textbf{25} & 15/57/\textbf{28} & 19/53/\textbf{28} \\
    \midrule
     \multirow{2}{*}{SQuALITY} & Broad (46\%) & \textbf{24}/55/21 & \textbf{28}/56/16 & \textbf{26}/58/16 \\
         & Specific (54\%) & 19/58/\textbf{23} & 14/57/\textbf{29} & 16/55/\textbf{29} \\
    \bottomrule
        
    \end{tabular}
    }
    \caption{\ours vs. \qontsum}
    \label{subtab:first_table_query_type}
  \end{subtable}
  \\
  \\
  \begin{subtable}{1\linewidth}
    \centering
    \scalebox{1}{
    \begin{tabular}{llcccc}
    \toprule
        \textbf{Dataset} & \textbf{Query type (\%)} & \textbf{Flu.} &\textbf{ Rel.} & \textbf{Faith.} \\ 
        \midrule
         \multirow{2}{*}{QMSum} & Broad (53\%) & 16/66/\textbf{18} & \textbf{41}/29/29 & \textbf{35}/32/32 \\
         & Specific (47\%) & 17/65/\textbf{18} & 20/38/\textbf{42} & 27/33/\textbf{40} \\
    \midrule
     \multirow{2}{*}{SQuALITY} & Broad (46\%) & \textbf{21}/58/18 & \textbf{35}/41/24 & \textbf{31}/41/\textbf{28} \\
         & Specific (54\%) & 18/60/\textbf{22} & 26/32/\textbf{42} & 21/47/\textbf{32} \\
    \bottomrule
        
    \end{tabular}
    
    }
    \caption{\ours vs. \socratic}
    \label{subtab:second_table_query_type}
  \end{subtable}
  \caption{Query type impact per model and model comparison with respect to query type. The reported numbers show the win/tie/lose \% of \ours against the baselines (i.e., \qontsum and \socratic), respectively.
  }
  \label{tab:query_type}
\end{table*}
 Table \ref{tab:human} reports the human evaluation scores over QMSum and SQuALITY datasets. As observed, the \ours model shows superior qualitative performance as compared to the \qontsum and \socratic baselines on both datasets. {To be more specific, the} \ours {model achieves a 5\% improvement in relevance and 4.3\% in faithfulness on the QMSum dataset, and a 2.8\% improvement in relevance and 2.4\% in faithfulness on the SQuALITY dataset. more pronounced in the relevance and faithfulness metrics, likely due to the }\ours model's focus on identifying segments that are more relevant to the query, prioritizing their relative importance. The close performance of the experimented systems over {fluency} is expected, given the extensive data the language model has encountered during pre-training to learn to generate coherent text. 
 
 {The inter-rater agreement scores are as follows: for QMSum, 51\%, 52\%, and 55\% and for SQuALITY, 51\%, 57\%, and 54\% across fluency, relevance, and faithfulness metrics, respectively, indicating a moderate level of consensus among evaluators.} While automatic improvements are numerically improved, our system still offers benefits in terms of qualitative (over \qontsum and \socratic) and training overhead (over \socratic) baselines, as mentioned earlier. {This assessment addresses our RQ1 on qualitative performance. }


\subsection{Query type impact} {We observed a potential relation between the system's qualitative performance and the nature of the query (i.e., query type)}. Specifically, we noticed that \textbf{broad queries} like \textit{``Summarize the discussion about price issues and target groups of remote control''} tend to have cover more gold segments to be answered as opposed to \textbf{specific queries} like \textit{``Why did the Marketing disagree with the Industrial Design when discussing the possible advanced techniques on the remote control?''}, targeting particular details within the long source. {It is important to note that in our study, the terms ``broad'' and ``specific'' are characterized by not only the breadth or specificity of the query itself, but also the number of segments needed to answer the query.} To explore this, we categorized the evaluation cases from each dataset based on their query type and compared the human-assigned scores to explore any potential links between the query type and the quality of the generated summaries.

Table \ref{tab:query_type} presents a comparison of the \ours system against \qontsum and \socratic systems, categorized by query types across two datasets. For broad queries, \ours outperforms \qontsum and \socratic, with notable win rates highlighted in bold; e.g., win rates of 37\% (QMSum), and 33\% (SQuALITY) in terms of relevance against \qontsum. However, with specific queries, our system's performance drops, often trailing the \qontsum and \socratic baselines, as evidenced by the high lose rates in bold; {e.g., 32\% (QMSum) and 34\% (SQuALITY) lose rates in relevance compared to \qontsum.} This trend, both highs and lows, is consistent across all qualitative metrics for both datasets. The differential performance of \ours vs. \qontsum and \socratic across query types can be attributed to the inherent granularity. In other words, {broad queries give} \ours {more room to maneuver since they cover a wide range of gold segments, available for ranking by the LTR component of our model.} However, specific queries are trickier; they focus on narrow details within narrow segments, where any slight oversight by the model in identifying salient segments leads to a less relevant summary. In the case of \socratic, the outperformance on specific queries can be attributed to its particular pre-training objective, where narrowed questions are generated for document's single sentences, and the language model is forced to learn to ask \& answer the generated questions. {Likewise, \qontsum excels in handling specific queries compared to broad queries, suggesting that its contrastive objective is more effective when there are fewer gold segments associated with the query, thereby enhancing the robustness of the objective.} This analysis addresses our RQ2. 

\begin{figure}[t]
    \centering
    \includegraphics[scale=0.29]{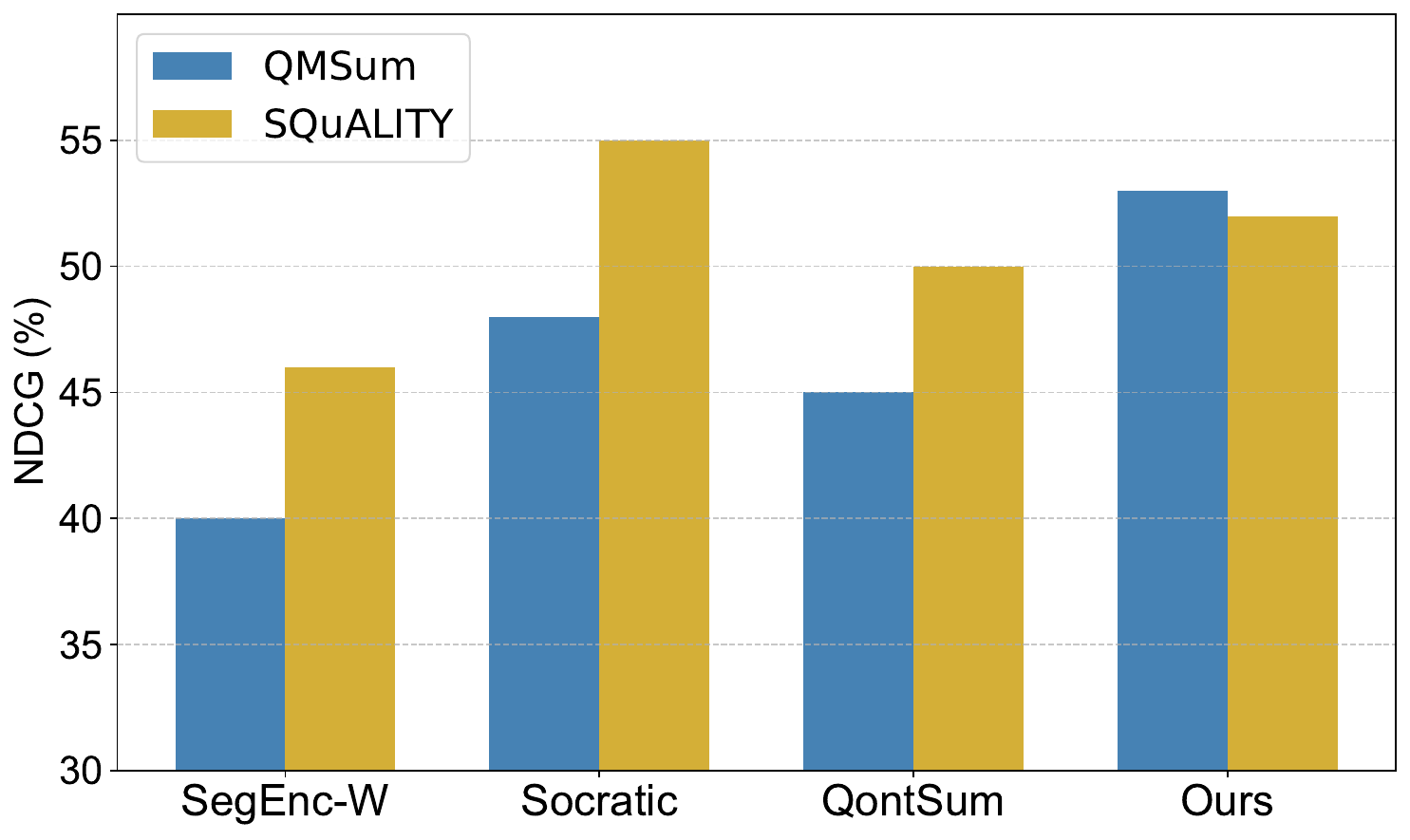}
    \caption{Segment retrieval performance of the models in terms of nDCG score. }
    \label{fig:seg}
\end{figure}

\begin{table*}[t]
\begin{minipage}[t]{.5\textwidth}
    \centering
    \begin{subtable}{\textwidth}
    \centering\scriptsize
    \scalebox{1.2}{
    \begin{tabular}{|p{6.8cm}|}\hline
    \textbf{Query:} Why did the Marketing disagree with the Industrial Design when discussing the possible advanced techniques on the remote control?
    \\
    \hline
    \textbf{Human:}
    When discussing adding several advanced techniques onto the remote control to make it more unique, the Industrial Design put forward to using the touchscreen. Notwithstanding the merits of the touchscreen, for instance, a touch screen would make the remote control easier and smaller,\textcolor{black}{\hl{the Marketing did not agree to do so just because in that case they would be beyond the budget.} \textit{[Written from the 14th segment]}} \\
    \hline
    \textbf{\ours generated:}
    \textcolor{black}{\hl{The Marketing believed that it would be too expensive to make a touch screen on the remote control.}}
     The Industrial Design believed that the strength of remote controls were most likely to fall down on the ground and get broken. Therefore, it would not be easy to make it fold open and look like a remote control with a touchscreen. \textit{[Written from the 15th segment]} \\
    \hline
    \end{tabular}}
    \end{subtable}
\end{minipage}%
\begin{minipage}[t]{.5\textwidth}
    \centering
    \begin{subtable}{\textwidth}
    \centering\scriptsize
    \scalebox{1.2}{
    \begin{tabular}{|p{4.5cm}|}\hline
    \textbf{Query:} What did User Interface think about user interface design of remote control?
    \\
    \hline
    \textbf{Human:}
    User Interface found two kinds of remote controls: the multi-functional one and the one easy to use. He emphasized on user-friendliness, but considering that the target people were less than forty years old, multi-function should also be taken into account. \textit{[Written from the 9th segment]} \\
    \hline
    \textbf{\ours generated:}
    User Interface thought that the remote control should be easy to use with not so many buttons, just a round button which can be pushed in four directions instead of a stick. It should be the same as in the cell phone, just light in the device that shines on all the buttons. \textit{[Written from the 9th segment]} \\
    \hline
    \end{tabular}}
    \end{subtable}
\end{minipage}
\caption{\small{Comparison between human and \ours generated summaries for given queries. Left: The model identifies relevant content (highlighted in yellow) from the 15th segment, which is marked in gold due to its 50\% overlap with the 14th segment, but also generates irrelevant information from the same 15th segment.} Right: The model finds the gold segment (segment 9) but picks up on less relevant parts of the segment.}
\label{fig:note}
\end{table*}
\subsection{Segment retrieval} 
{In order to assess the effectiveness of the summarization system in terms of lifting salient segments w.r.t their relative importance (i.e., ranking), we present a comparative analysis in Figure.}~\ref{fig:seg}. {To perform this analysis, we first rank the segments per summarization model, given their relative contribution (computed from decoder's attention over the segment tokens) at generating the summary. Subsequently, with the predicted ranked list of segments in hand, we calculate the Normalized Discounted Cumulative Gain (NDCG) score}~\cite{Wang2013ATA} {as follows:}

\begin{equation*}
    \text{DCG}_p = \sum_{i=1}^{p} \frac{2^{rel_i} - 1}{\log(i + 1)}
\end{equation*}

\begin{equation*}
    \text{nDCG}_p = \frac{\text{DCG}_p}{\text{IDCG}_p}
\end{equation*}
where $p$ is a particular ranking position, $rel_i$ is the relevance score (ranking label) of the segment at position $i$, and ${\text{IDCG}_p}$ is the ideal cumulative gain (i.e., when the segments are ranked given their gold importance). The relevance scores are obtained by greedily matching the system's ranked segments against the human-annotated important segments. As observed in Figure ~\ref{fig:seg}, our system consistently improves the ranking scores on QMSum and is comparable with the best-performing baseline (\socratic) on SQuALITY dataset. This analysis provides support for RQ3.

\section{Error Analysis}

Two sources of underperformance were identified {in response to our RQ4:}

\paragraph{\textbf{Imbalanced Labels.}} We discovered that in approximately 48\% of the underperformed cases, the model exhibited a tendency to misidentify gold segments when generating summaries. {Upon further investigation, we observed that these cases were commonly characterized by a label imbalance issue, wherein the number of gold segments was significantly smaller than non-gold segments. In such cases, the model selected segments that contained partially relevant information but were not the actual gold segments. As shown in the example within Table}~\ref{fig:note} {(left), while both human and} \ours{-generated summaries capture the \textit{budgetary concerns},} \ours adds unrelated information about \textit{remote control durability}. This finding sheds light on the challenge of identifying and ranking the gold segments within an imbalanced regime, which may be mitigated in future work through Transfer Learning from a larger dataset~\cite{ruder-etal-2019-transfer,Cao2019LearningID}.

\paragraph{\textbf{Segment Summarizer {Deficiency}. }}  {In approximately 39\% of the underperformed cases, } \ours faced challenges in extracting the most pertinent details from the identified gold segments. For instance, as illustrated in Table~\ref{fig:note} (right), both the human-written summary and the summary generated by \ours drew from the 9th segment (gold). The human summary provided a nuanced understanding of the topic, emphasizing both \textit{user-friendliness} and \textit{multi-functionality} for a \textit{specific age group}. Conversely, the \ours summary focused more on the \textit{physical attributes} of the \textit{remote control}, missing out on the \textit{multi-functionality aspect} and the \textit{target demographic}. This observed suboptimality could be attributed to the model's challenges in discerning sentential saliency within the segment which affects the relevancy of the summary. To address this, future work might consider hybrid approaches that combine methods for identifying salient sentences within the identified segments~\cite{pilault-etal-2020-extractive}. 


\section{Conclusion}


{Our method combines Learning-to-Rank with long-input QFS, ensuring content relevance via prioritization. The experimental results demonstrated that our proposed method matches or exceeds SOTA at reduced training costs. Human evaluations highlight improved relevance and faithfulness without compromising fluency. Further analysis suggests that the system outperforms on broad queries while lagging on specific ones, with errors linked to imbalanced labels and segment summarizer challenges.}


\section{Limitations}
While the proposed summarization system in our paper offers time-saving benefits, it still may produce outputs factually inconsistent with input documents or contain hallucinated information. Such discrepancies risk promoting online misinformation, especially when it is being used on the production scale. This challenge is common in abstractive summarization, necessitating rigorous research and cautious use to prevent false information spread.

\bibliography{latex/custom}

\end{document}